\documentclass[submission,copyright,creativecommons]{eptcs}
\providecommand{\event}{ICLP 2025} 

\ifpdf
  \usepackage{underscore}         
  \usepackage[T1]{fontenc}        
\else
  \usepackage{breakurl}           
\fi

\usepackage{amsmath}
\usepackage{graphicx}
\usepackage{multirow}

\usepackage{soul}
\usepackage{amssymb}
\usepackage{booktabs}
\usepackage{algorithm}
\usepackage[noend]{algpseudocode}
\usepackage{listings}
\usepackage{tikz}
\usetikzlibrary{arrows,shapes,automata,petri,positioning,calc}
\usepackage{listings}
\usepackage{bm}
\usepackage{url}

\usepackage[caption=false]{subfig}

\newcommand{\matr}[1]{\bm{#1}}  
\newtheorem{example}{Example}

\newtheorem{definition}{Definition}

\begin{document}

\title{Formally Explaining Decision Tree Models with \\Answer Set Programming}
\author{Akihiro Takemura
\institute{National Institute of Informatics, Tokyo, Japan}
\email{atakemura@nii.ac.jp}
\and
Masayuki Otani
\institute{Tokyo Institute of Technology, Tokyo, Japan}
\email{otani.m.af@m.titech.ac.jp}
\and
Katsumi Inoue
\institute{National Institute of Informatics, Tokyo, Japan}
\email{inoue@nii.ac.jp}
}
\def\titlerunning{Formally Explaining Decision Tree Models with Answer Set Programming}
\def\authorrunning{A. Takemura, M. Otani \& K. Inoue}

\maketitle

\begin{abstract}
Decision tree models, including random forests and gradient-boosted decision trees, are widely used in machine learning due to their high predictive performance. 
However, their complex structures often make them difficult to interpret, especially in safety-critical applications where model decisions require formal justification. 
Recent work has demonstrated that logical and abductive explanations can be derived through automated reasoning techniques. 
In this paper, we propose a method for generating various types of explanations, namely, sufficient, contrastive, majority, and tree-specific explanations, using Answer Set Programming (ASP). 
Compared to SAT-based approaches, our ASP-based method offers greater flexibility in encoding user preferences and supports enumeration of all possible explanations. 
We empirically evaluate the approach on a diverse set of datasets and demonstrate its effectiveness and limitations compared to existing methods.
\end{abstract}

\section{Introduction}\label{sec:introduction}
The widespread adoption of decision tree models in machine learning can be attributed to their high accuracy and computational efficiency. 
However, more complex ensemble variants, such as random forests and gradient-boosted trees, tend to operate as black-box models, rendering their internal decision-making processes opaque to users \cite{guidottiSurveyMethodsExplaining2018}. 
This lack of transparency is particularly problematic in high-stakes applications, such as healthcare, finance, and criminal justice, where trust in model decisions hinges on a clear understanding of how predictions are made.

Explainability, often referred to as eXplainable AI (XAI), seeks to bridge this gap by making model behavior understandable \cite{guidottiSurveyMethodsExplaining2018}. 
In practice, however, producing faithful explanations is challenging due to the complexity of modern machine learning models. 
Naive or heuristic-based techniques often fall short because they fail to capture the full logical structure underlying the model’s decisions \cite{rudinInterpretableMachineLearning2022a,marques-silvaDeliveringTrustworthyAI2022}.

Well-known post-hoc explanation methods such as LIME \cite{ribeiroWhyShouldTrust2016}, SHAP \cite{lundberg2017unified}, and Anchors \cite{ribeiroAnchorsHighprecisionModelagnostic2018} rely on heuristic approximations, and can yield explanations that are not guaranteed to align with the model’s true behavior \cite{narodytskaAssessingHeuristicMachine2019a}. 
In contrast, logic-based approaches offer rigorous guarantees and have been successfully applied to generate various types of explanations, including sufficient and contrastive ones \cite{izzaExplainingDecisionTrees2020,izzaExplainingRandomForests2021,ignatievAbductionBasedExplanationsMachine2019,audemardComputingAbductiveExplanations2023b,audemardContrastiveExplanationsTreeBased2023,audemardTradingComplexitySparsity2022,audemardExplanatoryPowerBoolean2022}. 
These logic-based approaches have been largely based on techniques of Boolean satisfiability (SAT). 
However, SAT-based methods often lack flexibility, especially when it comes to incorporating user-defined preferences or supporting complex reasoning tasks like explanation enumeration.

In this work, we propose an alternative approach using Answer Set Programming (ASP), a declarative logic programming paradigm well-suited for combinatorial reasoning. 
We show how ASP can be leveraged to encode and compute different types of explanations, including sufficient, contrastive, majority, and tree-specific explanations, for decision tree models. 
The high expressiveness of ASP allows us to concisely capture various explanation types and flexibly incorporate constraints, making it a promising tool for formal explainability.

Our main contributions are as follows:
\begin{enumerate}
    \item We develop ASP encodings for multiple explanation types, including sufficient, contrastive, majority, and tree-specific explanations.
    \item We implement and evaluate our approach across 20 datasets, comparing its performance against existing SAT-based methods.
    \item We analyze the ASP-based method's strengths and limitations, highlighting where it excels (majority explanation enumeration) and challenges (scalability to large forests).
\end{enumerate}

This paper is organized as follows: Section~\ref{sec:background} introduces the necessary background on decision tree models and ASP. 
Section~\ref{sec:expl_decision_tree} presents our ASP-based encodings for generating various types of explanations. 
Section~\ref{sec:experiments} details the experimental setup and results comparing our method with SAT-based approaches. 
Section~\ref{sec:relatedwork} discusses related work in the domain of explainable AI for decision tree models. 
Section~\ref{sec:conclusion} presents the conclusion.
Additionally, the Appendix provides worked examples and visual illustrations of our method applied to decision tree, random forest, and boosted tree models.

\section{Background}\label{sec:background}
This section introduces the necessary concepts for the rest of the paper.
We first review Boolean functions and their relationship to logical reasoning in the context of machine learning, and the structures of decision tree models. We then present a concise overview of ASP, the declarative framework used to encode and compute formal explanations in our approach.

\subsection{Boolean Functions}\label{sec:boolean_function}
Let \( X_n = \{x_1, \ldots, x_n\} \) be a set of Boolean variables.
A \textit{Boolean function} is a mapping $f: \{0,1\}^n \to \{0,1\}$, which assigns each Boolean assignment \( \matr{x} \in \{0,1\}^n \) a Boolean value.
We say \( \matr{x} \) is a \textit{positive instance} if \( f(\matr{x}) = 1 \), and a \textit{negative instance} if \( f(\matr{x}) = 0 \). 
A propositional formula over \(X_n\) is constructed using logical connectives: conjunction (\(\land\)), disjunction (\(\lor\)), and negation (\(\neg\)). 
A \textit{literal} \( L \) is either a variable \( x_i \) (positive literal) or its negation \( \neg x_i \) (negative literal). 
A \textit{term} is a conjunction of literals.

For an assignment \( \matr{z} \in \{0,1\}^n \), \(t_{\matr{z}} = \bigwedge_{i=1}^n x_i^{z_i}\) represents its corresponding term, where \( x_i^{0} = \neg x_i \) and \( x_i^{1} = x_i\). 
A term $t_1$ is a \textit{subterm} of a term $t_2$ if the literals appearing in $t_1$ appear in $t_2$.  
For example, $x_1 \land x_3$ is a subterm of  $x_1 \land \neg x_2 \land x_3$.
A term \( t \) \textit{covers} \( \matr{z} \) if \(t\) is a subterm of \(t_{\matr{z}}\), i.e., \(t\) is satisfied under \(\matr{z}\) \cite{audemardExplanatoryPowerBoolean2022}.
This captures the idea that \(t\) holds whenever the assignment \(\matr{z}\) satisfies all literals in \(t\).
A term \( t \) is an \textit{implicant} of a propositional formula \(f\) if \( t \models f \), i.e., every assignment \(\matr{z} \in \{0,1\}^n\) that satisfies \(t\) 
also satisfies \(f\). 
A term $t$ is a \textit{prime implicant} of $f$ 
if 
$t \models f$ and no proper subterm of \(t\) entails \(f\).

\subsection{Trees and Forests}\label{sec:tree_and_forest}
We consider a finite set of attributes (or features) \(\mathcal{A} = \{A_1, \dots, A_n\}\), where each attribute \(A_i\) takes values from a domain \(D_i\), which may be Boolean, categorical, or numerical. 
Decision tree learning algorithms create Boolean tests: threshold tests \((A_i \leq c)\) for numerical attributes and membership tests \((A_i \in S)\) for categorical attributes. 
Let \(L = \{L_1, \dots, L_n\}\) denote the Boolean literals from tree node tests, where each \(L_i\) corresponds to a Boolean variable from \(X_n\).
An instance \(\matr{z} = (v_1, \dots, v_n)\) with \(v_i \in D_i\) evaluates each Boolean variable \(x_i\) based on the tree node  tests. 
The corresponding term is \(t_{\matr{z}} = \bigwedge_{i=1}^n x_i^{z_i}\), where \(z_i = 1\) if the Boolean test for \(x_i\) evaluates to true and \(z_i = 0\) if it evaluates to false.


\paragraph{Decision Trees.} 
A \textit{decision tree} classifier over \(\mathcal{A}\) is a binary tree \(T\) where internal nodes test Boolean conditions and leaves are labeled with classes in \(\{0, 1\}\).
Tree traversal starts at the root and follows left branches for true conditions and right branches for false conditions.

\paragraph{Random Forests.} 
A \textit{random forest} \(RF = \{T_1, \dots, T_m\}\) is a set of decision trees $T_i$s, where its output is determined by majority vote \cite{breimanRandomForests2001}:
\begin{equation*}
    RF(\matr{x}) = 
\begin{cases}
1 & \text{if } \sum_{i=1}^m T_i(\matr{x}) > \frac{m}{2}, \\
0 & \text{otherwise}.
\end{cases}
\end{equation*}

\paragraph{Regression Trees and Forests.}
A \textit{regression tree} is similar to a decision tree, except leaves are labeled with real-valued outputs. 
A \textit{forest} $F = \{T_1, \dots, T_{m}\}$ consists of $m$ regression trees with aggregated weight $w(F, \matr{x}) = \sum_{k=1}^{m} w(T_k, \matr{x})$.

\paragraph{Boosted Trees.}
A \textit{boosted tree} $BT = \{T_1, \dots, T_m\}$ for binary classification is a sequence of trees  \cite{friedmanGreedyFunctionApproximation2001} (typically regression trees, as in gradient boosted trees such as XGBoost \cite{chenXGBoostScalableTree2016}) with prediction determined by the sign of cumulative weights.
\begin{equation*}
    BT(\matr{x}) = 
\begin{cases}
1 & \text{if } w(BT, \matr{x}) > 0, \\
0 & \text{if } w(BT, \matr{x}) < 0,
\end{cases}
\;\; \mathrm{where}\;\; w(BT, \matr{x}) = \sum_{k=1}^{m} w(T_k, \matr{x}).
\end{equation*}
We define the size of a forest \(F\), and a boosted tree \(BT\), as \(|F| = |BT| = \sum_{k=1}^{m} |T_k|\), where \(|T_k|\) is the number of nodes in the tree \(T_k\). 

\subsection{Answer Set Programming}
\textit{Answer Set Programming} (ASP) is a form of declarative programming oriented towards solving hard combinatorial problems.
It allows one to encode a problem as a set of logical rules, and uses a solver to compute its solutions, called \textit{answer sets}, based on the stable model semantics \cite{gelfondStableModelSemantics1988}.

A typical ASP rule is written as:
\begin{equation*}
    \mathrm{a_1} \ \texttt{:-} \ \mathrm{a_2},\ \dots,\ \mathrm{a_m},\ \texttt{not}\ \mathrm{a_{m+1}},\ \dots,\ \texttt{not}\ \mathrm{a_n}.
\end{equation*}
Here, each \(\mathrm{a_i}\) is a first-order atom, and \texttt{not} denotes \textit{default negation}. 
A rule with only the head \(\mathrm{a_1}\) is called a \textit{fact}, while one without a head is an \textit{integrity constraint}, used to eliminate undesired models.

Modern ASP systems, such as \textit{clingo} \cite{DBLP:journals/corr/GebserKKS14}, support expressive constructs that simplify problem modeling, for example:
\begin{itemize}
    \item Conditional literals of the form \(\{\mathrm{a}(\texttt{X}) : \mathrm{b}(\texttt{X})\}\), which represent sets of atoms \(\mathrm{a}(\texttt{X})\) for all values of \(\texttt{X}\) satisfying \(\mathrm{b}(\texttt{X})\).
    \item Cardinality constraints of the form \(s_1\ \{ \mathrm{a}(\texttt{X}) : \mathrm{b}(\texttt{X}) \}\ s_2\), which restrict the number of true literals in the set to be between \(s_1\) and \(s_2\).
    \item Aggregates such as \(\texttt{\#count}\) and \(\texttt{\#sum}\) allows one to count and sum the elements, respectively.
\end{itemize}

These constructs allow for compact and readable encodings of complex search spaces, making ASP particularly well-suited for formal explanations. 
Moreover, the \(\texttt{\#heuristic}\) directive provides solver-level preferences for atom selection.
For further details, see the \textit{clingo} user guide.\footnote{\url{https://github.com/potassco/guide/releases/}}

\section{Explaining Decision Tree Models}\label{sec:expl_decision_tree}
This section presents ASP-based encodings for extracting different types of explanations from decision tree models. 
We consider four explanation types (sufficient, contrastive, majority, and tree-specific), across three model types (decision trees, random forests, and gradient-boosted trees). 
Each explanation type is formally defined, and its corresponding ASP encoding is provided.

Our ASP encodings preserve the logical semantics of decision tree models through faithful representation of structure and decision flow, with integrity constraints ensuring only valid explanations.

Table~\ref{tab:supported_explanations} summarizes the explanation types supported for each model type in this paper.
\begin{table}[htbp]
    \centering
    \begin{tabular}{l cccc}
        \toprule
        Model            & Sufficient & Contrastive & Majority   & Tree-Specific \\
        \midrule
        Decision Tree    & \checkmark & \checkmark  &  n/a       & n/a \\
        Random Forest    & \checkmark & \checkmark  & \checkmark & n/a \\
        XGBoost          & -          &  -          &  n/a       & \checkmark \\
        \bottomrule
    \end{tabular}
    \caption{Explanation types supported by our ASP encodings for each model type.
    ``-'' indicates not implemented; "n/a" indicates not applicable.}
    \label{tab:supported_explanations}
\end{table}

\subsection{Decision Tree}
We begin with decision trees, where each model is translated into a set of ASP facts that represent its structure and prediction logic.
Each explanation type is defined in this subsection, along with its corresponding ASP encoding.
The predicates introduced by our preprocessing script are listed in Table~\ref{tab:dt_predicates}, where \texttt{node/3} and \texttt{pre\_class/1} are instance-dependent (truth values depend on the specific instance), while structural predicates like \texttt{leaf\_node/2} are instance-independent.
Across our encodings, including Random Forest and Boosted Trees, we use the \texttt{next\_node/2} predicate, which simulates path traversal from the root to a leaf.
By conditionally reconstructing the decision flow based on selected literals, it enables the solver to check whether an explanation preserves or alters the prediction, mirroring how decision trees arrive at their output.
See Appendix \ref{fig:appdx-dt} and Program \ref{pgm:appdx_dt_asp} for an example.
\begin{table}[htbp]
    \centering
\small{
    \begin{tabular}{l l}
    \toprule
      Predicate   & Meaning \\
      \midrule
      \texttt{node(X,L,B)}       & Node \texttt{X} checks whether literal \texttt{L} has truth value \texttt{B} in the current instance. \\
      \texttt{leaf\_node(X,C)}   & Node \texttt{X} is a leaf predicting class \texttt{C}. \\
      \texttt{left\_node(X,LX)}  & Node \texttt{X} has left child \texttt{LX}. \\
      \texttt{right\_node(X,RX)} & Node \texttt{X} has right child \texttt{RX}. \\
      \texttt{pre\_class(P)}     & The predicted class of the instance is \texttt{P}. \\
      \bottomrule
    \end{tabular}
}
    \caption{Predicates used to encode decision tree models as ASP facts.}
    \label{tab:dt_predicates}
\end{table}

\subsubsection{Sufficient Explanation from a Decision Tree}
\begin{definition}[Sufficient Explanation \cite{audemardExplanatoryPowerBoolean2022}]
  Let \( T \) be a decision tree over Boolean variables 
  \( X_n = \\ \{x_1, \dots, x_n\} \), let \( f \) be the Boolean function represented by \( T \), and let \( \matr{x} \in \{0,1\}^n \) be an instance with term \( t_{\matr{x}} \). 
  A \textit{sufficient explanation} for \( \matr{x} \) to be classified 1 (resp. 0) given \( f \) is a subterm \(t\) of \(t_{\matr{x}}\) such that \( t \) is a prime implicant of \( f \) (resp. \( \neg f \)).
\end{definition}
By definition, changing any literal in a prime implicant will change the output of the corresponding Boolean function.
Based on this, the sufficient explanation can be obtained by finding the maximal set of literals that can be changed without affecting the output, then taking the complement.
The rule \texttt{node(X,L,0..1)} (Line 2) allows conditional traversal in both directions, and the \texttt{invalid} predicate (Line 8) detects output changes.
\begin{lstlisting}[caption={Sufficient Explanation from a Decision Tree}{\vspace{4pt}},label={pgm:dt_suf}]
1 {selected_literal(L):node(X,L,B)}.
node(X,L,0..1) :- selected_literal(L),node(X,L,B).
next_node(LX) :- node(0,L,1),left_node(0,LX).
next_node(RX) :- node(0,L,0),right_node(0,RX).
next_node(LX) :- next_node(X),node(X,L,1),left_node(X,LX).
next_node(RX) :- next_node(X),node(X,L,0),right_node(X,RX).
class(C):-next_node(X),leaf_node(X,C).
invalid :- class(0),class(1).
:- invalid.   %% remove solutions that change the prediction
#heuristic selected_literal(L). [1,true]   %% prefer including more literals
#show selected_literal/1.
\end{lstlisting}
This encoding will find the maximal set of \texttt{selected\_literal} by utilizing the \texttt{\#heuristic} directive in \textit{clingo} which will prioritize assigning \texttt{true} to \texttt{selected\_literal}. See Appendix Example \ref{ex:appdx_dt_suf} for an example.

\subsubsection{Contrastive Explanation from a Decision Tree}
Unlike sufficient explanations, we are looking for a minimal set of literals that will change the output of the Boolean function.
\begin{definition}[Contrastive Explanation \cite{audemardExplanatoryPowerBoolean2022}]\label{def:dt_con}
  Let \( T \) be a decision tree over Boolean variables \( X_n = \\ \{x_1, \dots, x_n\} \), let \( f \) be the Boolean function represented by \( T \), and let \( \matr{x} \in \{0,1\}^n \) be an instance with term \( t_{\matr{x}} \). 
  A \textit{contrastive explanation} for \(\matr{x}\) to be classified 1 (resp. 0) given \(f\) is a subterm \(t\) of \(t_{\matr{x}}\) such that removing the literals in \(t\) from \(t_{\matr{x}}\) yields a term that is not an implicant of \(f\) (resp. \(\neg f\)), and for every literal \(l \in t\), removing all literals in \(t\) except \(l\) from \(t_{\matr{x}}\) does not satisfy this condition.
\end{definition}
Thus, we allow for this change in the prediction by introducing the \texttt{valid} predicate (Line 8) which must not be false according to the integrity constraint on the next line.
\begin{lstlisting}[caption={Contrastive Explanation from a Decision Tree}{\vspace{4pt}},label={pgm:dt_con}]
1 {selected_literal(L):node(X,L,B)}.
node(X,L,0..1) :- selected_literal(L),node(X,L,B).
next_node(LX) :- node(0,L,1),left_node(0,LX).
next_node(RX) :- node(0,L,0),right_node(0,RX).
next_node(LX) :- next_node(X),node(X,L,1),left_node(X,LX).
next_node(RX) :- next_node(X),node(X,L,0),right_node(X,RX).
class(C) :- next_node(X),leaf_node(X,C).
valid :- class(0),class(1).
:- not valid.   %% the output must change
#heuristic selected_literal(L). [1,false]  %% prefer including fewer literals
#show selected_literal/1.
\end{lstlisting}
This encoding will find the minimal set of \texttt{selected\_literal} by utilizing the \texttt{\#heuristic} directive in \textit{clingo} which will prioritize assigning \texttt{false} to \texttt{selected\_literal}. See Appendix Example \ref{ex:appdx_dt_con} for an example.


\subsubsection{Worked Example: Sufficient Explanations from a Decision Tree}
\begin{figure}[htbp]
    \centering
    \begin{tikzpicture}
        \tikzstyle{level 1}=[level distance=1.5cm, sibling distance=4cm]
        \tikzstyle{level 2}=[level distance=1.5cm, sibling distance=2.5cm]

        \tikzset{decision/.style={
            draw=black, rectangle, 
            inner sep=2mm, text centered} 
        }
        \tikzset{leaf/.style={
            draw=black, circle,
            inner sep=1mm, text centered
        }}

        \node [decision, label=above:{\scriptsize Node 0 (\(L_0\))}] (root) {\(x_1 \leq 2\)}
            child {
                node [decision, label=above:{\scriptsize Node 1 (\(L_1\))}] {\(x_2 \leq 3\)}
                    child {
                        node [leaf, label=above:{\scriptsize Node 2}] {1}
                        edge from parent node[left, above] {T}
                    }
                    child {
                        node [leaf, label=above:{\scriptsize Node 3}] {0}
                        edge from parent node[right, above] {F}
                    }
                edge from parent node[left, above] {T}
            }
            child {
                node [decision, label=above:{\scriptsize Node 4 (\(L_2\))}] {\(x_3 \leq 1\)}
                    child {
                        node [leaf, label=above:{\scriptsize Node 5}] {1}
                        edge from parent node[left, above] {T}
                    }
                    child {
                        node [leaf, label=above:{\scriptsize Node 6}] {0}
                        edge from parent node[right, above] {F}
                    }
                edge from parent node[right, above] {F}
            };
    \end{tikzpicture}
    \caption{Example decision tree for Boolean attributes.}
    \label{fig:dt_example_1}
\end{figure}
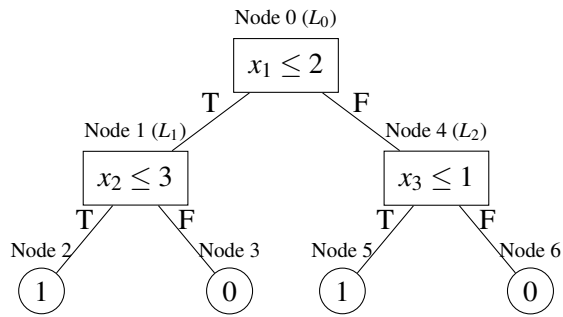

Consider the decision tree in Figure~\ref{fig:dt_example_1} with instance $\matr{x} = \{x_1 = 1, x_2 = 1, x_3 = 2\}$.

The preprocessing converts this tree to Boolean literals: $(x_1 \leq 2)$ becomes literal $L_0$ (true), $(x_2 \leq 3)$ becomes literal $L_1$ (true), and $(x_3 \leq 1)$ becomes literal $L_2$ (false). 
The instance follows the left-left path, reaching leaf node 2 with prediction 1.
The ASP encoding represents this instance as a set of facts:
\begin{lstlisting}
pre_class(1). node(0,0,1). node(1,1,1). node(4,2,0).
leaf_node(2,1). leaf_node(3,0). leaf_node(5,1). leaf_node(6,0).
left_node(0,1). right_node(0,4). % ... additional facts
\end{lstlisting}

For sufficient explanation, Program~\ref{pgm:dt_suf} produces \texttt{selected\_literal(2)}, meaning literals $L_0$ and $L_1$ form the sufficient explanation $\{x_1 \leq 2, x_2 \leq 3\}$.

\subsection{Random Forest}
Random forests are ensembles of decision trees where the final prediction is determined by majority voting. 
We describe how sufficient, contrastive, and majority explanations can be derived from random forests using ASP.
Each tree is encoded as described previously, with additional predicates to track selected literals and tree agreement.
Table~\ref{tab:rf_predicates} lists the predicates used to represent forest structure and control explanation logic.
\begin{table}[htbp]
    \centering
\small{
    \begin{tabular}{l l}
    \toprule
      Predicate   & Meaning \\
      \midrule
      \texttt{node(T,X,L,B)}       & In tree \texttt{T}, node \texttt{X} checks whether literal \texttt{L} has truth value \texttt{B} for this instance. \\
      \texttt{class(T,C)} & The tree \texttt{T} predicts class \texttt{C} for this instance.\\
      \texttt{feature(L,B)} & The literal \texttt{L} has truth value \texttt{B} for this instance.\\
      \texttt{new\_node(T,X,L,B)}  & Copy of the node with truth value possibly flipped.\\
      \texttt{next\_node(T,X)}     & Reachable node under possibly changed truth value.\\
      \texttt{leaf\_node(T,X,C)}   & In tree \texttt{T}, node \texttt{X} is a leaf predicting class \texttt{C}. \\
      \texttt{left\_node(T,X,LX)}  & Left child of node \texttt{X} in tree \texttt{T} is \texttt{LX}. \\
      \texttt{right\_node(T,X,RX)} & Right child of node \texttt{X} in tree \texttt{T} is \texttt{RX}. \\
      \texttt{pre\_forest(P)}      & Predicted class of the forest for \(\matr{x}\) is \texttt{P}. \\
      \texttt{tree\_threshold(TH)} & Threshold for majority in sufficient explanations. \\
      \texttt{con\_tree\_threshold(TH)} & Threshold for majority in contrastive explanations. \\
      \texttt{majo\_tree\_threshold(TH)} & Threshold for majority explanations (e.g., \( \lceil m/2 \rceil \)). \\
      \bottomrule
    \end{tabular}
}
    \caption{Predicates used to encode random forest models as ASP facts.}
    \label{tab:rf_predicates}
\end{table}

\subsubsection{Sufficient Explanation from a Random Forest}
\begin{definition}[Sufficient Explanation \cite{audemardTradingComplexitySparsity2022}]\label{def:rf_suf}
  Let \( RF = \{T_1, \dots, T_m\} \) be a random forest over Boolean variables \( X_n = \{x_1, \dots, x_n\} \), and let \( f \) be the Boolean function represented by \( RF \).
  Let \( \matr{x} \in \{0,1\}^n \) be an instance with associated term \( t_{\matr{x}} \).
  A \textit{sufficient explanation} for \(\matr{x}\) to be classified 1 (resp. 0) given \(f\) is a subterm \(t\) of \(t_{\matr{x}}\) such that \(t\) is a prime implicant of \(f\) (resp. \(\neg f\)).
\end{definition}

A \textit{minimal unsatisfiable subset} (MUS) is a set of atoms that enforce the absence of an answer set.
For random forests, sufficient explanations require fixing a minimal set of atoms that block all counterfactuals.
We adopt a two-stage approach using \textit{WASP} \cite{alvianoWASPNativeASP2013} with unsatisfiable core extraction:
Stage 1 searches for counterfactuals that change the forest's prediction; Stage 2 computes a MUS of literals that must be fixed to prevent this change.

The first encoding below defines the logical structure of the forest and detects whether a change in the instance leads to a different prediction.
The second encoding uses WASP’s \texttt{--mus} mode to extract a minimal set of fixed literals.
The following encoding computes sufficient explanations by identifying a maximal set of literals such that enough trees preserve the prediction.
\begin{lstlisting}[caption={Sufficient Explanation from a Random Forest stage 1}{\vspace{4pt}},label={pgm:rf_suf_1}]
node(T,X,L) :- node(T,X,L,_).
new_node(T,X,L,0) :- change(L),node(T,X,L),feature(L,1).
new_node(T,X,L,1) :- change(L),node(T,X,L),feature(L,0).
new_node(T,X,L,B) :- node(T,X,L),feature(L,B),not change(L).
next_node(T,LX) :- new_node(T,0,L,1),left_node(T,0,LX).
next_node(T,RX) :- new_node(T,0,L,0),right_node(T,0,RX).
next_node(T,LX) :- next_node(T,X),new_node(T,X,L,1),left_node(T,X,LX).
next_node(T,RX) :- next_node(T,X),new_node(T,X,L,0),right_node(T,X,RX).
class(T,C) :- next_node(T,X),leaf_node(T,X,C).
invalid_tree(T) :- class(T,C),pre_forest(FC),C!=FC.
forest_changed :- VT=#count{T:invalid_tree(T)},tree_threshold(TH),VT>TH.
:- not forest_changed.
\end{lstlisting}
The second stage computes a MUS of literals that must be fixed to prevent the forest's prediction from changing, based on the counterfactual structure defined in Stage 1. 
This is done by invoking WASP with \texttt{--mus} mode, which extracts a minimal unsatisfiable core over the guessed subset of fixed literals.
\begin{lstlisting}[caption={Sufficient Explanation from a Random Forest stage 2}{\vspace{4pt}},label={pgm:rf_suf_2}]
feature(I,B). ...  %% list of features involved in the prediction of forest
fix_lit(I). ...  %% guessing atoms for MUS extraction (see Appendix)
\end{lstlisting}

Program~\ref{pgm:rf_suf_2} shows the high-level structure. 
The \texttt{feature/2} facts represent the original instance, while \texttt{fix\_lit/1} guesses which literals must retain their original value. 
This method allows us to extract a minimal set of literals that are \textit{necessary} to preserve the original prediction, effectively computing a sufficient explanation. See Appendix Example \ref{ex:appdx_rf_suf} for an example.

\subsubsection{Contrastive Explanation from a Random Forest}
\begin{definition}[Contrastive Explanation \cite{audemardTradingComplexitySparsity2022}]\label{def:rf_con}
  Let \( RF = \{T_1, \dots, T_m\} \) be a random forest over Boolean variables \( X_n = \{x_1, \dots, x_n\} \), let \( f \) be the Boolean function represented by \( RF \), and let  \( \matr{x} \in \{0,1\}^n \) be an instance with associated term \( t_{\matr{x}} \).
  A \textit{contrastive explanation} for \(\matr{x}\) to be classified 1 (resp. 0) given \(f\) is a subterm \(t\) of \(t_{\matr{x}}\) such that removing the literals in \(t\) from \(t_{\matr{x}}\) yields a term that is not an implicant of \(f\) (resp. \(\neg f\)), and for every literal \(l \in t\), removing \(l\) from \(t\) does not satisfy this condition.
\end{definition}
This encoding searches for a minimal set of literals whose removal flips the random forest output. 
The \texttt{selected\_literal/1} predicate determines which literals change, and \texttt{con\_tree\_threshold/1} enforces the required number of disagreeing trees.
\begin{lstlisting}[caption={Contrastive Explanation from a Random Forest}{\vspace{4pt}},label={pgm:rf_con}]
1 {selected_literal(L):node(T,X,L,B)}.
new_node(T,X,L,0) :- selected_literal(L),node(T,X,L,1).
new_node(T,X,L,1) :- selected_literal(L),node(T,X,L,0).
new_node(T,X,L,B) :- node(T,X,L,B),not selected_literal(L).
next_node(T,LX) :- new_node(T,0,L,1),left_node(T,0,LX).
next_node(T,RX) :- new_node(T,0,L,0),right_node(T,0,RX).
next_node(T,LX) :- next_node(T,X),new_node(T,X,L,1),left_node(T,X,LX).
next_node(T,RX) :- next_node(T,X),new_node(T,X,L,0),right_node(T,X,RX).
class(T,C):-next_node(T,X),leaf_node(T,X,C).
valid_tree(T) :- class(T,C),pre_forest(FC),C!=FC.
valid :- VT = #count{T : valid_tree(T)},con_tree_threshold(TH),VT>TH.
:- not valid.
#heuristic selected_literal(L). [1,false]
#show selected_literal/1.
\end{lstlisting}

\subsubsection{Majority Explanation from a Random Forest}
\begin{definition}[Majority Explanation \cite{audemardTradingComplexitySparsity2022}]\label{def:rf_maj}
  Let \( RF = \{T_1, \dots, T_m\} \) be a random forest over Boolean variables \( X_n = \{x_1, \dots, x_n\} \), let \( f \) be the Boolean function represented by \( RF \) and let  \( \matr{x} \in \{0,1\}^n \) be an instance with associated term \( t_{\matr{x}} \).
  A \textit{majority explanation} for \(\matr{x}\) to be classified 1 (resp. 0) given \(f\) is a subterm \(t\) of \(t_{\matr{x}}\) such that \(t\) is an implicant of at least \(\lfloor m/2 \rfloor + 1\) trees \(T_i\) if \(RF(\matr{x}) = 1\) (resp. \(\neg T_i\) if \(RF(\matr{x}) = 0\)), and for every literal \(l \in t\), removing \(l\) from \(t\) does not satisfy this condition.
\end{definition}
This encoding searches for a maximal term \(t\) that can be flipped without violating the majority vote by counting trees that agree with the forest's original prediction. 
The explanation must retain enough agreement across individual trees to ensure that the forest prediction remains consistent.

\begin{lstlisting}[caption={Majority Explanation from a Random Forest}{\vspace{4pt}},label={pgm:rf_maj}]
1 {selected_literal(L):node(T,X,L,B)}.
node(T,X,L,0..1) :- selected_literal(L),node(T,X,L,B).
next_node(T,LX) :- node(T,0,L,1),left_node(T,0,LX).
next_node(T,RX) :- node(T,0,L,0),right_node(T,0,RX).
next_node(T,LX) :- next_node(T,X),node(T,X,L,1),left_node(T,X,LX).
next_node(T,RX) :- next_node(T,X),node(T,X,L,0),right_node(T,X,RX).
class(T,C) :- next_node(T,X),leaf_node(T,X,C).
invalid_tree(T) :- class(T,C),pre_forest(FC),C!=FC.
valid :- VT = #count{T : invalid_tree(T)},majo_tree_threshold(TH),VT<TH.
:- not valid.
#heuristic selected_literal(L). [1,true]
#show selected_literal/1.
\end{lstlisting}
This encoding uses \texttt{majo\_tree\_threshold/1} to specify the minimum number of agreeing trees required to preserve the original forest prediction. 
The \texttt{\#heuristic} directive encourages literal inclusion, producing a maximal term that is filtered to identify minimal majority explanations.
Since the encoding finds maximal sets of literals that can be flipped, the actual majority explanation is the complement: the literals that must remain fixed to preserve the majority vote.
See Appendix Example \ref{ex:appdx_rf_maj} for an example.

\subsection{Boosted Trees}
We now turn to boosted tree models, such as XGBoost \cite{chenXGBoostScalableTree2016} and LightGBM \cite{keLightGBMHighlyEfficient2017}, where prediction is made by aggregating outputs from regression trees.
Let $BT = \{T_1, \dots, T_m\}$ ($1 \leq i \leq m$) composed of \(m\) trees denote a boosted tree model, where each leaf contributes a real-valued weight and the final prediction is determined by the sign of the sum of weights.
Table~\ref{tab:xgb_predicates} lists the predicates used to encode boosted trees as ASP facts.
\begin{table}[htbp]
    \centering
\small{
    \begin{tabular}{l l}
    \toprule
      Predicate   & Meaning \\
    \midrule
    \texttt{node(T,X,L,B)} & In tree, \texttt{T}, node \texttt{X} checks whether literal \texttt{L} has truth value \texttt{B} in the current instance. \\
    \texttt{leaf\_node(T,X,W)} & Node \texttt{X} in tree \texttt{T} is a leaf with weight \texttt{W}. \\
    \texttt{left\_node(T,X,LX)} & Left child of node \texttt{X} in tree \texttt{T} is \texttt{LX}. \\
    \texttt{right\_node(T,X,RX)} & Right child of node \texttt{X} in tree \texttt{T} is \texttt{RX}. \\
    \texttt{pre\_forest(P)} & Original prediction of \texttt{X} by the forest is \(\texttt{P} \in \{0,1\}\). \\
    \texttt{weight(T,W)}  & Leaf weight reached under current selection in tree \texttt{T}. \\
    \texttt{best\_weight(T,BW)}	& Maximum weight reachable in tree \texttt{T}. \\
    \texttt{worst\_weight(T,WW)} & Minimum weight reachable in tree \texttt{T}. \\
    \bottomrule
    \end{tabular}
}
    \caption{Predicates used to encode boosted tree models as ASP facts.}
    \label{tab:xgb_predicates}
\end{table}

\subsubsection{Tree-specific explanation from a Boosted Tree}
Tree-specific explanations for boosted trees identify minimal sets of literals that guarantee a prediction under best-case or worst-case scenarios.
\begin{definition}[Best and Worst Instance \cite{audemardComputingAbductiveExplanations2023b}]
  Let \( BT = \{T_1, \dots, T_m\} \) be a boosted tree over Boolean variables \( X_n = \{x_1, \dots, x_n\} \), and let \( \matr{x}, \matr{x}', \matr{x}'' \in \{0,1\}^n \) be instances with corresponding terms \( t_{\matr{x}}, t_{\matr{x}'}, t_{\matr{x}''} \), and assume \( BT(\matr{x}) = BT(\matr{x}') = BT(\matr{x}'') \).
  \begin{itemize}
      \item The \textit{best instance} extending \( t \) is any \( \matr{x}' \) such that \(t\) is a subterm of \(t_{\matr{x}'}\), \( BT(\matr{x}') = BT(\matr{x}) \), and \( \matr{x}' = \arg\max_{\matr{x}'' : \, t \text{ is a subterm of } t_{\matr{x}''}} (\{w(BT, \matr{x}'') \}) \).
      \( w_{\uparrow}(t,BT) \) is the weight of any best instance, i.e., \( w_{\uparrow}(t,BT) = \max_{\matr{x}'' : \, t \text{ is a subterm of } t_{\matr{x}''}} w(BT, \matr{x}'')\). 
      \item The \textit{worst instance} extending \( t \) is any \( \matr{x}' \) such that \(t\) is a subterm of \(t_{\matr{x}'}\), \( BT(\matr{x}') = BT(\matr{x}) \), and \( \matr{x}' = \arg\min_{\matr{x}'' : \, t \text{ is a subterm of } t_{\matr{x}''}} (\{w(BT, \matr{x}'') \})\).       
      \( w_{\downarrow}(t,BT) \) is the weight of any worst instance, i.e., \( w_{\downarrow}(t,BT) = \min_{\matr{x}'' : \, t \text{ is a subterm of } t_{\matr{x}''}} w(BT, \matr{x}'')\). 
  \end{itemize}
\end{definition}
The best (resp. worst) instances represent the most (resp. least) favorable scenarios, allowing reasoning about prediction guarantees under pessimistic conditions.
\begin{definition} [Tree-Specific Explanation \cite{audemardComputingAbductiveExplanations2023b}]\label{def:bt_ts}
Let $BT = \{T_1, \dots, T_m\}$ ($1 \leq i \leq m$) be a boosted tree, $\matr{x}$ be the input instance. 
Formally, let $w_{\downarrow}(t, T_i)$ denote the minimum (worst-case) score that tree $T_i$ can produce under any instance extending $t$, and $w_{\uparrow}(t, T_i)$ the maximum (best-case) score. 
Then, $t$ is a \textit{tree-specific explanation} of instance $\matr{x}$ if it satisfies:
\begin{itemize}
  \item When $BT(\matr{x})=1$: \(t\) is a subterm of \(t_{\matr{x}}\), and $\sum_{i=1}^m w_{\downarrow}(t, T_i) > 0$ and no proper subterm of $t$ satisfies the latter condition.
  \item When $BT(\matr{x})=0$: \(t\) is a subterm of \(t_{\matr{x}}\), and $\sum_{i=1}^m w_{\uparrow}(t, T_i) \leq 0$ and no proper subterm of $t$ satisfies the latter condition.
\end{itemize}
\end{definition}
\begin{lstlisting}[caption={Tree-specific Explanation from a Boosted Tree}{\vspace{4pt}},label={pgm:bt_ts}]
1 {selected_literal(L) : node(T,X,L,B) }.
node(T,X,L,0..1) :- node(T,X,L,B), not selected_literal(L).
next_node(T,LX) :- node(T,0,L,1), left_node(T,0,LX).
next_node(T,RX) :- node(T,0,L,0), right_node(T,0,RX).
next_node(T,LX) :- next_node(T,X), node(T,X,L,1), left_node(T,X,LX).
next_node(T,RX) :- next_node(T,X), node(T,X,L,0), right_node(T,X,RX).
weight(T,W) :- next_node(T,X), leaf_node(T,X,W).
best_weight(T,BW) :- weight(T, _), BW = #max{W:weight(T,W)}.
worst_weight(T,WW) :- weight(T, _), WW = #min{W:weight(T,W)}.
valid :- SW = #sum{BW:best_weight(_,BW)}, SW<=0, pre_forest(0).
valid :- SW = #sum{WW:worst_weight(_,WW)}, SW>0, pre_forest(1).
:- not valid.
#heuristic selected_literal(L). [1,false]
#show selected_literal/1.
\end{lstlisting}

This encoding preserves decisions under pessimistic aggregation using \texttt{\#min} and \texttt{\#max} aggregates (Lines 8 and 9) for modelling tree extremal behavior and \texttt{\#sum} aggregates (Lines 10 and 11) for forest-level decisions.
See Appendix Example \ref{ex:appdx_bt_ts} for an example.

\section{Experiments}\label{sec:experiments}
To evaluate our approach, we conducted experiments on 20 publicly available datasets from OpenML, UCI, and Kaggle.
We used 100 test instances per dataset sampled from a 20\% test split, to ensure that evaluation reflects generalization performance on unseen data. 
We set a 100-second timeout per instance, a maximum depth of 6, and 100 estimators for ensemble models.

We compare our ASP-based method against PyXAI \cite{audemardExplanatoryPowerBoolean2022}, a SAT-based framework for formal explanations.
LIME \cite{ribeiroWhyShouldTrust2016}, SHAP \cite{lundberg2017unified} and Anchors \cite{ribeiroAnchorsHighprecisionModelagnostic2018} are excluded from this comparison, since they provide heuristic approximations without formal guarantees \cite{ignatievTrustableExplainableAI2020}.
Experiments were conducted on Ubuntu 22.04 with Intel Core i9-9900K and 64 GB RAM.

Table~\ref{tab:exp_success} shows success rates for computing explanations within the timeout, distinguishing between computing \textit{all} explanations versus \textit{one} explanation.
A successful computation means that at least one explanation was produced for the \textit{one} case, and that all explanations were enumerated for the \textit{all} case.
Our ASP approach achieved perfect success on decision trees and was competitive on random forests when computing a single explanation.
However, ASP struggled when enumerating explanations for RFs, especially for contrastive and majority explanations. 
In contrast, for enumerating explanations for RFs, PyXAI showed better performance in enumerating contrastive explanations, but ASP was better in enumerating majority explanations.

\begin{table}[htbp]
    \centering
\small{
    \begin{tabular}{lllrr}
    \toprule
    Type & Num. explanations & Model & PyXAI completed (\%) & ASP completed (\%) \\
    \midrule
    Contrastive & All & DT & 100.0 & 100.0 \\
    Contrastive & All & RF & 42.6 & 6.4 \\
    Contrastive & One & DT & 100.0 & 100.0 \\
    Contrastive & One & RF & 60.8 & 100.0 \\
    Majority & All & RF & 0.4 & 6.4 \\
    Majority & One & RF & 100.0 & 98.7 \\
    Sufficient & All & DT & 100.0 & 100.0 \\
    Sufficient & One & DT & 100.0 & 100.0 \\
    Sufficient & One & RF & 100.0 & 57.7 \\
    Tree Specific & One & XGB & 100.0 & 20.8 \\
    \bottomrule
    \end{tabular}
}
    \caption{Success rates of computing explanations across explanation types and models.}
    \label{tab:exp_success}
\end{table}

\begin{table}[ht]
    \centering
\footnotesize{
\begin{tabular}{llrrrrrrrr}
\toprule
\multirow{2}{*}{} & \multirow{2}{*}{dataset} & 
\multicolumn{2}{c}{Con \#N} & \multicolumn{2}{c}{Maj \#N} & 
\multicolumn{2}{c}{Con \#L} & \multicolumn{2}{c}{Maj \#L} \\
\cmidrule(lr){3-4} \cmidrule(lr){5-6} \cmidrule(lr){7-8} \cmidrule(lr){9-10}
& & PyXAI & ASP & PyXAI & ASP & PyXAI & ASP & PyXAI & ASP \\
\midrule
1 & ad\_data & 19.7 & - & - & 133277.5 & 1.7 & - & - & 181.1 \\
2 & adult & 11.8 & - & - & 72112.0 & 2.2 & - & - & 77.3 \\
3 & arcene & 179.0 & 179.0 & - & - & 1.0 & 1.0 & - & - \\
4 & australia & 7.9 & - & - & - & 1.8 & - & - & - \\
5 & banknote & 8.0 & 31.0 & - & - & 1.7 & 1.0 & - & - \\
6 & breast & 8.3 & 10.0 & 27.7 & 48202.8 & 2.7 & 1.0 & 33.8 & 30.6 \\
7 & christine & 2.7 & 157.0 & - & - & 2.4 & 1.0 & - & - \\
8 & compas & 3.9 & 3.8 & - & 22085.6 & 1.5 & 1.5 & - & 13.9 \\
9 & credit\_german & 7.5 & - & - & 1392.0 & 2.6 & - & - & 82.6 \\
10 & dexter & 50.7 & 172.8 & - & - & 2.2 & 1.0 & - & - \\
11 & diabetes & 11.7 & - & - & - & 1.7 & - & - & - \\
12 & divorce & 20.6 & - & - & - & 4.4 & - & - & - \\
13 & gina\_agnostic & 7.2 & - & - & - & 2.3 & - & - & - \\
14 & heart & 6.7 & - & - & 5768.0 & 2.0 & - & - & 81.1 \\
15 & indian\_liver\_patient & 24.5 & - & - & - & 1.5 & - & - & - \\
16 & qsar & 9.9 & - & - & - & 2.0 & - & - & - \\
17 & spambase & 1.7 & - & - & - & 1.2 & - & - & - \\
18 & startup & 14.0 & 476.1 & - & - & 2.1 & 1.0 & - & - \\
19 & titanic & 13.9 & 208.0 & - & 127778.0 & 2.5 & 1.0 & - & 212.9 \\
20 & wine & 6.0 & - & - & - & 1.0 & - & - & - \\
\bottomrule
\end{tabular}
}
    \caption{Random Forest results: average number (\#N) and length (\#L) of contrastive and majority explanations. Dash (-) indicates timeouts (100s). Results for completed instances only (all explanations).}
    \label{tab:exp_rf_data}
\end{table}

Table~\ref{tab:exp_rf_data} compares average number of explanations and lengths for Random Forest models, showing results only for completed instances.
Note that averages are computed only over successful instances, and PyXAI and ASP methods may succeed on different subsets.

ASP was able to enumerate more contrastive and majority explanations than PyXAI on the subset of instances where it completed execution within the timeout (\#N columns). 
This trend is particularly evident in the case of majority explanations, where ASP outperformed PyXAI significantly on several datasets (e.g., \textit{startup}, \textit{titanic}, \textit{breast}), despite a lower overall completion rate.
However, ASP faces scalability challenges for complex models and exhaustive enumeration under resource constraints.

\section{Related Work}\label{sec:relatedwork}
The problem of explaining decision tree-based models has received significant attention, particularly through logical and symbolic methods.
Audemard et al. \cite{audemardExplanatoryPowerBoolean2022} proposed a SAT-based framework for generating formal explanations from decision trees by translating the model into CNF, supporting \textit{direct}, \textit{sufficient}, and \textit{contrastive} that correspond to the explanation notions we adopt in this paper. 

Izza et al. \cite{izzaExplainingDecisionTrees2020} addressed redundancy in path-based explanations by extracting abductive explanations as literal subsets, offering more concise explanations than path tracing.

Several SAT-based extensions have targeted ensemble models. 
Audemard et al. extended their framework to compute \textit{majority explanations} \cite{audemardTradingComplexitySparsity2022} and \textit{contrastive explanations} \cite{audemardContrastiveExplanationsTreeBased2023} for random forests, demonstrating feasibility while acknowledging computational challenges in explanation enumeration.

Izza et al. \cite{izzaExplainingRandomForests2021} proposed an efficient SAT-based method to generate prime implicant (PI) explanations for random forests. 
Their results showed improved performance over heuristic methods such as Anchors \cite{ribeiroAnchorsHighprecisionModelagnostic2018}, highlighting the advantages of logical methods in producing faithful explanations.

Takemura et al. \cite{takemuraGeneratingGlobalLocal2024} proposed an ASP-based method to derive rule-based explanations from decision tree models. 
Their approach focused on extracting global and local rules through pattern mining from the structure of decision and ensemble trees. 
However, their explanations lack the logical guarantees provided by our approach.

Beyond tree models, Dhurandhar et al. \cite{dhurandharExplanationsBasedMissing2018} developed the Contrastive Explanation Method (CEM) for deep neural networks, introducing the contrastive paradigm and emphasizing explanations aligned with human intuition through the presence or absence of signals.

In contrast to these works, our ASP-based approach provides a unified, extensible framework for generating multiple types of formal explanations for decision tree models and ensembles. 
By leveraging ASP's model enumeration capabilities, we enable various explanation types within a single encoding environment, improving flexibility and simplifying integration with domain-specific constraints.

\section{Conclusion}\label{sec:conclusion}
This work presented an ASP-based approach for generating formal explanations for decision tree models. 
We implemented encodings for sufficient, contrastive, majority, and tree-specific explanations, demonstrating that ASP can reproduce and sometimes outperform SAT-based methods, particularly for individual majority and contrastive explanations on random forests.

ASP's enumeration capabilities offer practical advantages when generating multiple explanations by exploring the search space in a single solver call. 
However, scalability challenges remain for large models, where long computation time limits practical applicability. 
%
The primary bottlenecks are: (1) exponential memory growth during grounding when instantiating all possible combinations, and (2) the inefficient two-stage process requiring two distinct ASP systems.

Future work will focus on improving runtime performance through more efficient encodings, pruning, better heuristics, and optimization, while extending to boosted tree explanations and broader comparative analysis.

\subsection*{Acknowledgements}
This work has been supported by JSPS KAKENHI Grant Numbers JP21H04905, JP25K03190 and JST CREST Grant Number JPMJCR22D3, Japan.

\nocite{*}
\bibliographystyle{eptcs}
\bibliography{ref}


\newpage
\appendix
\section{Illustrative Examples of Explanations}
This appendix provides visual examples to illustrate how sufficient, contrastive, majority and tree-specific explanations can be derived from decision tree models. 
Each figure is accompanied by a brief explanation of what the model structure and literals represent.

\subsection{Decision Tree}
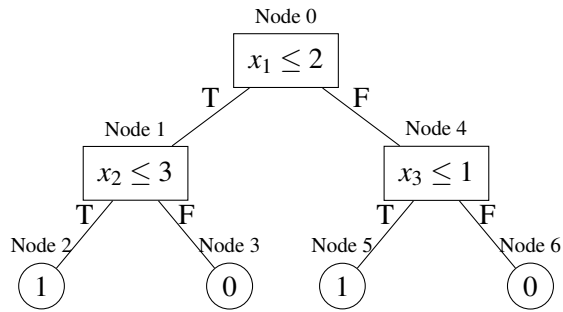
\begin{figure}[htbp]
    \centering
    \begin{tikzpicture}
        \tikzstyle{level 1}=[level distance=1.5cm, sibling distance=4cm]
        \tikzstyle{level 2}=[level distance=1.5cm, sibling distance=2.5cm]

        \tikzset{decision/.style={
            draw=black, rectangle, 
            inner sep=2mm, text centered} 
        }
        \tikzset{leaf/.style={
            draw=black, circle,
            inner sep=1mm, text centered
        }}

        \node [decision, label=above:{\scriptsize Node 0}] (root) {\(x_1 \leq 2\)}
            child {
                node [decision, label=above:{\scriptsize Node 1}] {\(x_2 \leq 3\)}
                    child {
                        node [leaf, label=above:{\scriptsize Node 2}] {1}
                        edge from parent node[left, above] {T}
                    }
                    child {
                        node [leaf, label=above:{\scriptsize Node 3}] {0}
                        edge from parent node[right, above] {F}
                    }
                edge from parent node[left, above] {T}
            }
            child {
                node [decision, label=above:{\scriptsize Node 4}] {\(x_3 \leq 1\)}
                    child {
                        node [leaf, label=above:{\scriptsize Node 5}] {1}
                        edge from parent node[left, above] {T}
                    }
                    child {
                        node [leaf, label=above:{\scriptsize Node 6}] {0}
                        edge from parent node[right, above] {F}
                    }
                edge from parent node[right, above] {F}
            };
    \end{tikzpicture}
    \caption{A decision tree with three internal nodes.}
    \label{fig:appdx-dt}
\end{figure}

Let the input instance be \(\matr{x} = \{x_1 = 1, x_2 = 1, x_3 = 2\}\). 
This instance satisfies \( (x_1 \leq 2) \) and \( (x_2 \leq 3) \), so the tree follows the left-left path and predicts \(1\).
The tree is encoded into a set of ASP facts, where we use the depth-first indexing starting at 0.
For example, \texttt{node(0,1,0)} corresponds to \((x_1 \leq 2)\) being satisfied, and \texttt{node(1,1,1)} corresponds to \((x_2 \leq 3)\) being satisfied, etc.

\begin{lstlisting}[caption={Decision Tree: Instance}{\vspace{4pt}},label={pgm:appdx_dt_asp}]
%% node status
node(0,0,1). node(1,1,1). node(4,1,0).

%% tree structure
leaf_node(2,1). leaf_node(3,0). leaf_node(5,1). leaf_node(6,0).
left_node(0,1). right_node(0,4). left_node(1,2). right_node(1,3).
left_node(4,5). right_node(4,6).
\end{lstlisting}

\begin{example}[Sufficient Explanation]\label{ex:appdx_dt_suf}
    Executing the Program~\ref{pgm:dt_suf} with this instance produces the answer set containing \texttt{selected\_literal(2)}, which is the complement of the sufficient explanation.
    Thus, the sufficient explanation for this case is: \texttt{selected\_literal(0)} and \texttt{selected\_literal(1)}. 
    These correspond to the literals \(\{x_1 \leq 2, x_2 \leq 3\}\), where changing either literal causes the prediction to flip, making the term a sufficient explanation.
\end{example}

\begin{example}[Contrastive Explanation]\label{ex:appdx_dt_con}
    Executing the program \ref{pgm:dt_con} yields \texttt{selected\_literal(1)}, which corresponds to \(\{x_2 \leq 3\}\).
    Flipping this condition, i.e., \(x_2 > 3\), causes the output to change from 1 to 0: this is a contrastive explanation.
\end{example}

\newpage

\subsection{Random Forest}
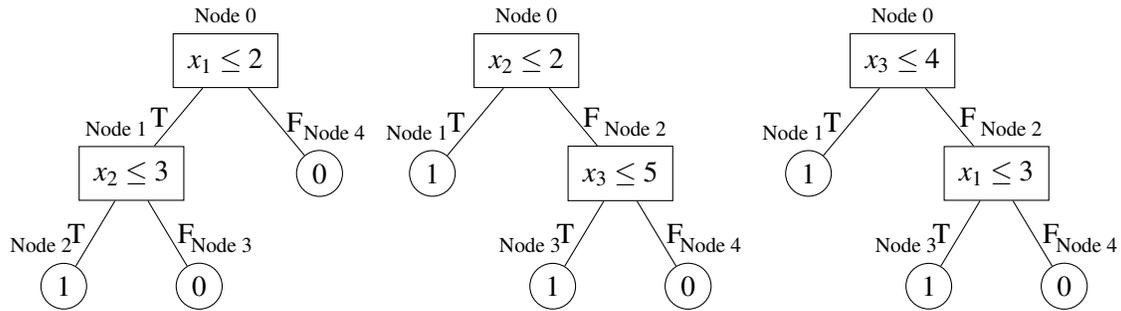
\begin{figure}[htbp]
    \centering
    \begin{tikzpicture}
    \tikzstyle{level 1}=[level distance=1.5cm, sibling distance=2.5cm]
    \tikzstyle{level 2}=[level distance=1.5cm, sibling distance=1.8cm]
    \tikzset{decision/.style={draw=black, rectangle, inner sep=2mm, text centered}}
    \tikzset{leaf/.style={draw=black, circle, inner sep=1mm, text centered}}

    \node [decision, label=above:{\scriptsize Node 0}] (t1root) at (0,0) {\(x_1 \leq 2\)}
        child {
            node [decision, label={[xshift=-2mm]above:{\scriptsize Node 1}}] {\(x_2 \leq 3\)}
                child { node [leaf, label={[xshift=-3mm]above:{\scriptsize Node 2}}] {1} edge from parent node[left] {T} }
                child { node [leaf, label={[xshift=3mm]above:{\scriptsize Node 3}}] {0} edge from parent node[right] {F} }
            edge from parent node[left] {T}
        }
        child { node [leaf, label={[xshift=2mm]above:{\scriptsize Node 4}}] {0} edge from parent node[right] {F} };

    \node [decision, label=above:{\scriptsize Node 0}] (t2root) at (4,0) {\(x_2 \leq 2\)}
        child {
            node [leaf, label={[xshift=-2mm]above:{\scriptsize Node 1}}] {1}
            edge from parent node[left] {T}
        }
        child {
            node [decision, label={[xshift=2mm]above:{\scriptsize Node 2}}] {\(x_3 \leq 5\)}
                child { node [leaf, label={[xshift=-3mm]above:{\scriptsize Node 3}}] {1} edge from parent node[left] {T} }
                child { node [leaf, label={[xshift=3mm]above:{\scriptsize Node 4}}] {0} edge from parent node[right] {F} }
            edge from parent node[right] {F}
        };

    \node [decision, label=above:{\scriptsize Node 0}] (t3root) at (9,0) {\(x_3 \leq 4\)}
        child {
            node [leaf, label={[xshift=-2mm]above:{\scriptsize Node 1}}] {1}
            edge from parent node[left] {T}
        }
        child {
            node [decision, label={[xshift=2mm]above:{\scriptsize Node 2}}] {\(x_1 \leq 3\)}
                child { node [leaf, label={[xshift=-3mm]above:{\scriptsize Node 3}}] {1} edge from parent node[left] {T} }
                child { node [leaf, label={[xshift=3mm]above:{\scriptsize Node 4}}] {0} edge from parent node[right] {F} }
            edge from parent node[right] {F}
        };

    \end{tikzpicture}
    \caption{A random forest with three trees.}
    \label{fig:appdx-rf-3tree}
\end{figure}

Let us consider the instance \(\matr{x} = \{x_1 = 1, x_2 = 2, x_3 = 6\}\), for which all three trees predict class 1.
This random forest and the instance results in the following encoding.
\begin{lstlisting}[caption={Random Forest: Instance}{\vspace{4pt}},label={pgm:appdx_rf_asp}]
%% instance and node status
feature(1,1).
feature(2,1).
feature(3,1).
feature(4,0).
feature(5,0).
feature(6,1).

%% forest prediction and thresholds
pre_forest(1). con_tree_threshold(1). majo_tree_threshold(2).

%% tree 1
node(1,0,1,1). node(1,1,2,1).
leaf_node(1,2,1). leaf_node(1,3,0). leaf_node(1,4,0).
left_node(1,0,1). right_node(1,0,4). left_node(1,1,2). right_node(1,1,3).

%% tree 2
node(2,0,3,1). node(2,2,4,0).
leaf_node(2,1,1). leaf_node(2,3,1). leaf_node(2,4,0).
left_node(2,0,1). right_node(2,0,2). left_node(2,2,3). right_node(2,2,4).

%% tree 3
node(3,0,5,0). node(3,2,6,1).
leaf_node(3,1,1). leaf_node(3,3,1). leaf_node(3,4,0). 
left_node(3,0,1). right_node(3,0,2). left_node(3,2,3). right_node(3,2,4).
\end{lstlisting}

\begin{lstlisting}[caption={Random Forest Sufficient Explanation: Encoding}{\vspace{4pt}},label={pgm:appdx_rf_asp_suf_enc}]

%% Prepend this line to Program 3 and 9, and save it as enc.lp
%% Generator for MUS extraction. Note that 1..6 is forest-dependent.
{change(L) : not fix_lit(L),L=1..6}.  

%% Program 3
node(T,X,L) :- node(T,X,L,_).
new_node(T,X,L,0) :- change(L),node(T,X,L),feature(L,1).
new_node(T,X,L,1) :- change(L),node(T,X,L),feature(L,0).
...

%% Program 9
%% instance and node status
feature(1,1).
feature(2,1).
...
\end{lstlisting}

\begin{lstlisting}[caption={Random Forest Sufficient Explanation: Instance}{\vspace{4pt}},label={pgm:appdx_rf_asp_suf_ins}]
%% For MUS extraction with WASP, save this as ins.lp
fix_lit(1).
fix_lit(2).
fix_lit(3).
fix_lit(4).
fix_lit(5).
fix_lit(6).
\end{lstlisting}

\begin{lstlisting}[caption={Random Forest Sufficient Explanation: Command}{\vspace{4pt}},label={pgm:appdx_rf_asp_suf_command},breaklines=true]
python wasp_rewriter.py enc.lp ins.lp | clingo --output=smodels | ./wasp --mus=__debug__ -n0
\end{lstlisting}

\begin{example}[Sufficient Explanation]\label{ex:appdx_rf_suf}
    Executing the command in Program \ref{pgm:appdx_rf_asp_suf_command} with WASP outputs \\\texttt{fix\_lit(3), fix\_lit(6)}, which corresponds to \(\{x_2 \leq 2, x_1 \leq 3\}\).
    Fixing these literals ensures that enough trees preserve the original prediction.
\end{example}

\begin{example}[Contrastive Explanation]\label{ex:appdx_rf_con}
    Executing the program \ref{pgm:rf_con} outputs \\
    \texttt{selected\_literal(1), selected\_literal(3)}, which corresponds to \(\{x_1 \leq 2, x_2 \leq 2\}\).
    If both of these literals are flipped, the majority of the trees change their prediction, and the forest's prediction will change.
\end{example}

\begin{example}[Majority Explanation]\label{ex:appdx_rf_maj}
    Executing the program \ref{pgm:rf_maj} outputs \\
    \texttt{selected\_literal(3), selected\_literal(4), selected\_literal(5)}, \\
    which, when subtracted from the set of literals, results in \\
    \texttt{selected\_literal(1), selected\_literal(2), selected\_literal(6)}.
    These correspond to \\ \(\{ x_1 \leq 2, x_2 \leq 3, x_1 \leq 3 \}\), which ensure that at least two trees maintain the prediction.
\end{example}

\newpage

\subsection{Boosted Trees}
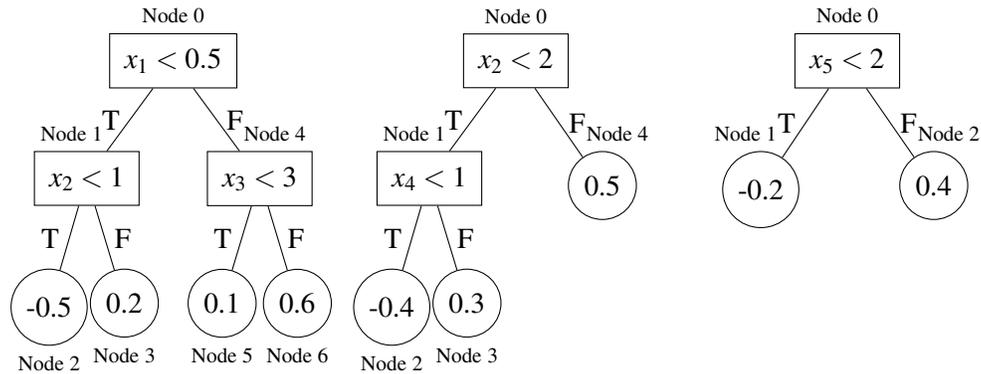
\begin{figure}[htbp]
\centering
\begin{tikzpicture}
\tikzstyle{level 1}=[level distance=1.2cm, sibling distance=2.3cm]
\tikzstyle{level 2}=[level distance=1.2cm, sibling distance=1.0cm]
\tikzstyle{level 3}=[level distance=1.2cm, sibling distance=0.5cm]
\tikzstyle{decision}=[draw=black, rectangle, anchor=north, inner sep=2mm, text centered]
\tikzstyle{leaf}=[draw=black, circle, anchor=north, text centered, minimum size=0.6cm]

\node [decision, label=above:{\scriptsize Node 0}] (t1root) at (0,0) {$x_1 < 0.5$}
  child {
    node [decision, label={[xshift=-2mm]above:{\scriptsize Node 1}}] {$x_2 < 1$}
      child {
        node [leaf, label={[]below:{\scriptsize Node 2}}] {-0.5}
        edge from parent node[left] {T}
      }
      child {
        node [leaf, label={[]below:{\scriptsize Node 3}}] {0.2}
        edge from parent node[right] {F}
      }
    edge from parent node[left] {T}
  }
  child {
    node [decision, label={[xshift=2mm]above:{\scriptsize Node 4}}] {$x_3 < 3$}
      child {
        node [leaf, label={[]below:{\scriptsize Node 5}}] {0.1}
        edge from parent node[left] {T}
      }
      child {
        node [leaf, label={[]below:{\scriptsize Node 6}}] {0.6}
        edge from parent node[right] {F}
      }
    edge from parent node[right] {F}
  };

\node [decision, right=3cm of t1root, label=above:{\scriptsize Node 0}] (t2root) {$x_2 < 2$}
  child {
    node [decision, label={[xshift=-2mm]above:{\scriptsize Node 1}}] {$x_4 < 1$}
      child {
        node [leaf, label={[]below:{\scriptsize Node 2}}] {-0.4}
        edge from parent node[left] {T}
      }
      child {
        node [leaf, label={[]below:{\scriptsize Node 3}}] {0.3}
        edge from parent node[right] {F}
      }
    edge from parent node[left] {T}
  }
  child {
    node [leaf, label={[xshift=2mm]above:{\scriptsize Node 4}}] {0.5}
    edge from parent node[right] {F}
  };

\node [decision, right=3cm of t2root, label=above:{\scriptsize Node 0}] (t3root) {$x_5 < 2$}
  child {
    node [leaf, label={[xshift=-2mm]above:{\scriptsize Node 1}}] {-0.2}
    edge from parent node[left] {T}
  }
  child {
    node [leaf, label={[xshift=2mm]above:{\scriptsize Node 2}}] {0.4}
    edge from parent node[right] {F}
  };

\end{tikzpicture}
\caption{Boosted tree model with three regression trees.}
\label{fig:appdx-boosted-tree}
\end{figure}

Consider the input instance \( \matr{x} = \{x_1 = 1,\ x_2 = 2,\ x_3 = 3,\ x_4 = 4,\ x_5 = 5\} \).
The boosted tree is encoded into ASP facts as follows:
\begin{lstlisting}[caption={Boosted Tree: Instance}{\vspace{4pt}},label={pgm:appdx_xgb_asp}]
pre_forest(1).
%% tree 1
node(0,0,1,0). node(0,1,2,1). node(0,4,3,0).
leaf_node(0,2,-500). leaf_node(0,3,200). leaf_node(0,5,100). leaf_node(0,6,600).
left_node(0,0,1). right_node(0,0,4).
left_node(0,1,2). right_node(0,1,3).
left_node(0,4,5). right_node(0,4,6).

%% tree 2
node(1,0,4,0). node(1,1,5,0).
leaf_node(1,2,-400). leaf_node(1,3,300). leaf_node(1,4,500).
left_node(1,0,1). right_node(1,0,4).
left_node(1,1,2). right_node(1,1,3).

%% tree 3
node(2,0,6,0). leaf_node(2,1,-200). leaf_node(2,2,400).
left_node(2,0,1). right_node(2,0,2).
\end{lstlisting}

\begin{example}[Tree-Specific Explanation]\label{ex:appdx_bt_ts}
    Executing the program \ref{pgm:bt_ts} outputs \texttt{selected\_literal(5), selected\_literal(6)}, which corresponds to \(\{x_4 < 1, x_5 < 2\}\).
    Fixing these literals ensure that the sum of the worst-case instances remain greater than 0.
\end{example}

\end{document}